\documentclass[11pt,a4paper]{article}
\usepackage[hyperref]{acl2018}
\usepackage{times}
\usepackage{latexsym}
\usepackage{multirow}
\usepackage{multicol}
\usepackage{graphicx}
\usepackage{tabularx}
\usepackage{subcaption}
\usepackage{makecell}
\usepackage{amssymb}
\usepackage{multirow}
\usepackage{amsmath}
\usepackage{enumitem}

\usepackage{url}

\DeclareCaptionFont{10pt}{\fontsize{10pt}{12pt}\selectfont}
\aclfinalcopy 

\title{Cold-Start Aware User and Product Attention\\for Sentiment Classification}

\author{Reinald Kim Amplayo \and
  Jihyeok Kim \and
  Sua Sung \and
  Seung-won Hwang \\
  Yonsei University \\
  Seoul, South Korea \\
  {\tt \{rktamplayo, zizi1532, dormouse, seungwonh\}@yonsei.ac.kr} \\
}

\date{}

\begin{document}
\maketitle
\begin{abstract}
  The use of user/product information in sentiment analysis is important, especially for \textit{cold-start} users/products, whose number of reviews are very limited. However, current models do not deal with the cold-start problem which is typical in review websites.
  In this paper, we present Hybrid Contextualized Sentiment Classifier (HCSC), which contains two modules:
  (1) a fast word encoder that returns word vectors embedded with short and long range dependency features; and
  (2) Cold-Start Aware Attention (CSAA), an attention mechanism that considers the existence of cold-start problem when attentively pooling the encoded word vectors.
  HCSC introduces \textbf{shared} vectors that are constructed from similar users/products, and are used when the original \textbf{distinct} vectors do not have sufficient information (i.e. cold-start). This is decided by a frequency-guided selective gate vector.
  Our experiments show that in terms of RMSE, HCSC performs significantly better when compared with on famous datasets, despite having less complexity, and thus can be trained much faster. More importantly, our model performs significantly better than previous models when the training data is sparse and has cold-start problems.
\end{abstract}

\section{Introduction}

Sentiment classification is the fundamental task of sentiment analysis \cite{pang2002thumbs}, where we are to classify the sentiment of a given text.
It is widely used on online review websites as they contain huge amounts of review data that can be classified a sentiment.
In these websites, a sentiment is usually represented as an intensity (e.g. 4 out of 5).
The reviews are written by users who have bought a product.
Recently, sentiment analysis research has focused on personalization \cite{zhang2015incorporating} to recommend product to users, and vise versa.

To this end, many have used user and product information not only to develop personalization but also to improve the performance of the classification model \cite{tang2015learning}. Indeed, these information are important in two ways. First, some expressions are user-specific for a certain sentiment intensity. For example, the phrase ``\textit{very salty}'' may have different sentiments for a person who likes salty food and a person who likes otherwise. This is also apparent in terms of products. Second, these additional contexts help mitigate data sparsity and cold-start problems. Cold-start is a problem when the model cannot draw useful information from users/products where data is insufficient. User and product information can help by introducing a frequent user/product with similar attributes to the cold-start user/product.

\begin{figure}
    \centering
    \includegraphics[width=0.47\textwidth]{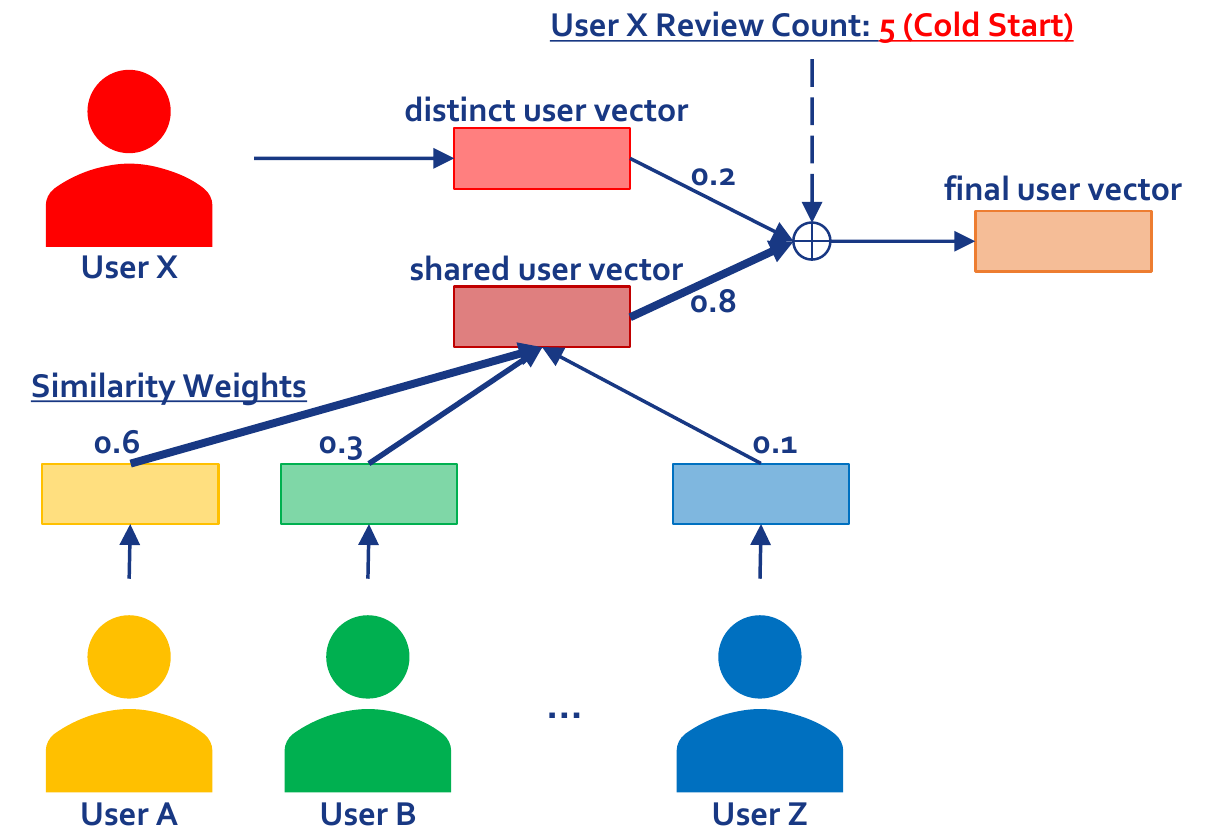}
    \caption{Conceptual schema of HCSC applied to users. The same idea can be applied to products.}
    \label{fig:intro}
\end{figure}

Thanks to the promising results of deep neural networks to the sentiment classification task \cite{glorot2011domain,tang2014coooolll}, more recent models incorporate user and product information to convolutional neural networks \cite{tang2015learning} and deep memory networks \cite{dou2017capturing}, and have shown significant improvements. The current state-of-the-art model, NSC \cite{chen2016neural}, introduced an attention mechanism called UPA which is based on user and product information
and applied this to a hierarchical LSTM.
The main problem with current models is that they use user and product information naively as an ordinary additional context, not considering the possible existence of cold-start problems.
This makes NSC more problematic than helpful in reality since majority of the users in review websites have very few number of reviews.

To this end, we propose the idea shown in Figure \ref{fig:intro}. It can be described as follows: If the model does not have enough information to create a user/product vector, then we use a vector computed from other user/product vectors that are similar.
We introduce a new model called Hybrid Contextualized Sentiment Classifier (HCSC), which consists of two modules.
First, we build a fast yet effective word encoder that accepts word vectors and outputs new encoded vectors that are contextualized with short- and long-range contexts.
Second, we combine these vectors into one pooled vector through a novel attention mechanism called Cold-Start Aware Attention (CSAA). The CSAA mechanism has three components: (a) a user/product-specific \textbf{distinct vector} derived from the original user/product information of the review, (b) a user/product-specific \textbf{shared vector} derived from other users/products, and (c) a frequency-guided \textbf{selective gate} which decides which vector to use.
Multiple experiments are conducted with the following results:
In the original non-sparse datasets, our model performs significantly better than the previous state-of-the-art, NSC, in terms of RMSE, despite being less complex.
In the sparse datasets, HCSC performs significantly better than previous competing models.

\section{Related work}
\label{sec:rel}

\begin{figure*}[t]
    \centering
    \includegraphics[width=0.97\textwidth]{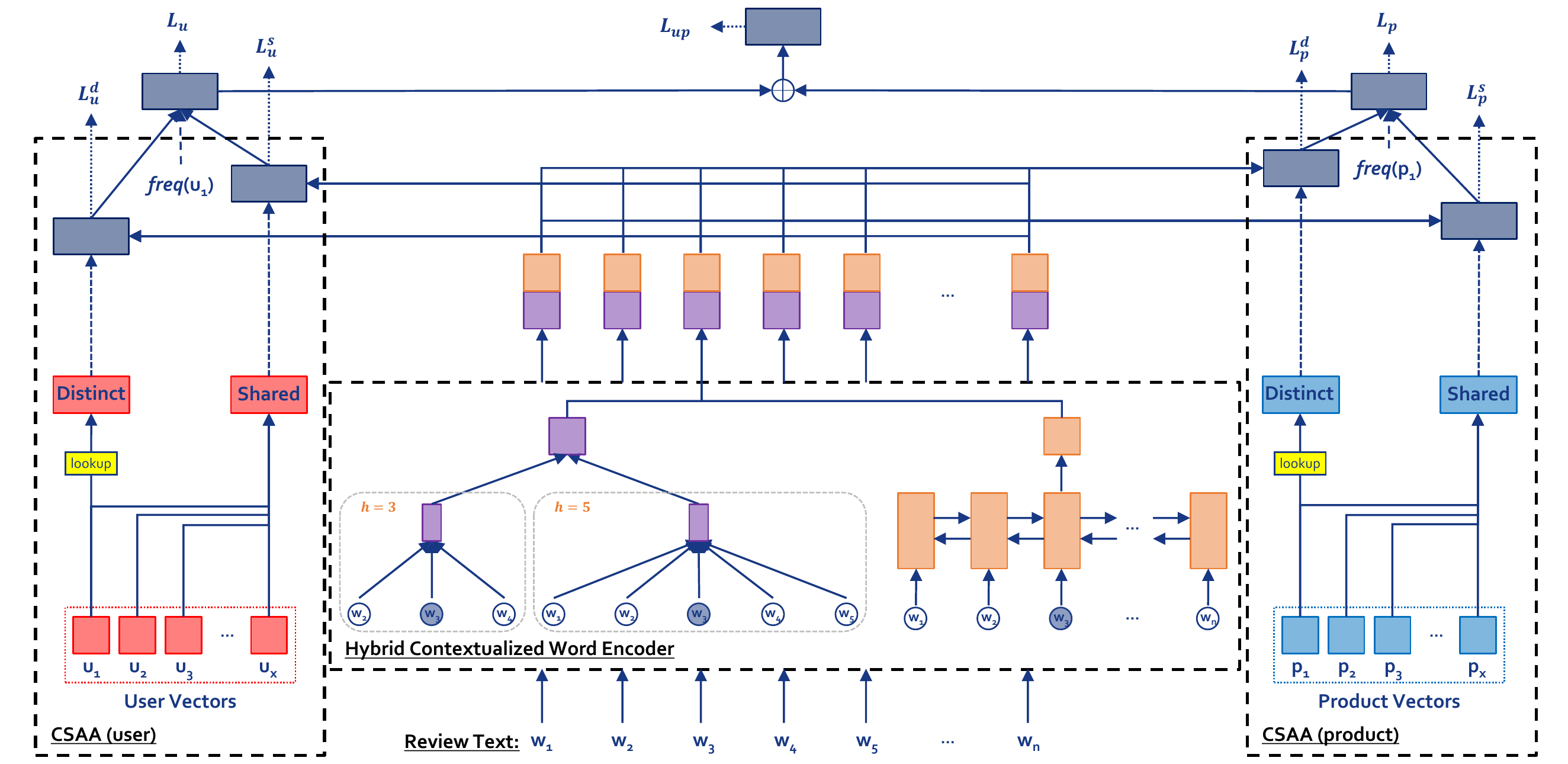}
    \caption{Full architecture of HCSC, which consists of the Hybrid Contextualized Word Encoder (middle), and user-specific (left) and product-specific (right) Cold-Start Aware Attention (CSAA).}
    \label{fig:model}
\end{figure*}

Previous studies have shown that using additional contexts for sentiment classification helps improve the performance of the classifier. We survey several competing baseline models that use user and product information and other models using other kinds of additional context.

\paragraph{Baselines: Models with user and product information}

User and product information are helpful to improve the performance of a sentiment classifier. This argument was verified by \citet{tang2015learning} through the observation at the consistency between user/product information and the sentiments and expressions found in the text. Listed below are the following models that employ user and product information:

\begin{itemize}
    \setlength\itemsep{-5pt}
    \item \textbf{JMARS} \cite{diao2014jointly} jointly models the aspects, ratings, and sentiments of a review while considering the user and product information using collaborative filtering and topic modeling techniques.
    \item \textbf{UPNN} \cite{tang2015learning} uses a CNN-based classifier and extends it to incorporate user- and product-specific text preference matrix in the word level which modifies the word meaning.
    \item \textbf{TLFM+PRC} \cite{song2017recommendation} is a text-driven latent factor model that unifies user- and product-specific latent factor models represented using the consistency assumption by \citet{tang2015learning}.
    \item \textbf{UPDMN} \cite{dou2017capturing} uses an LSTM classifier as the document encoder and modifies the encoded vector using a deep memory network with other documents of the user/product as the memory.
    \item \textbf{TUPCNN} \cite{chen2016learning} extends the CNN-based classifier by adding temporal user and product embeddings, which are obtained from a sequential model and learned through the temporal order of reviews.
    \item \textbf{NSC} \cite{chen2016neural} is the current state-of-the-art model that utilizes a hierarchical LSTM model \cite{yang2016hierarchical} and incorporates user and product information in the attention mechanism.
\end{itemize}

\paragraph{Models with other additional contexts}
Other additional contexts used previously are spatial \cite{yang2017identifying} and temporal \cite{fukuhara2007understanding} features which help contextualize the sentiment based on the location where and the time when the text is written. Inferred contexts were also used as additional contexts for sentiment classifiers, such as latent topics \cite{lin2009joint} and aspects \cite{jo2011aspect} from a topic model, argumentation features \cite{wachsmuth2015sentiment}, and more recently, latent review clusters \cite{amplayo2017aspect}. These additional contexts were especially useful when data is sparse, i.e. number of instances is small or there exists cold-start entities.

Our model differs from the baseline models mainly because we consider the possible existence of the data sparsity problem. Through this, we are able to construct more effective models that are comparably powerful yet more efficient complexity-wise than the state-of-the-art, and are better when the training data is sparse. Ultimately, our goal is to demonstrate that, similar to other additional contexts, user and product information can be used to effectively mitigate the problem caused by cold-start users and products.

\section{Our model}

In this section, we present our model, \textbf{Hybrid Contextualized Sentiment Classifier (HCSC)}\footnote{The data and code used in this paper are available here: \url{https://github.com/rktamplayo/HCSC}.} which consists of a fast hybrid contextualized word encoder and an attention mechanism called Cold-Start Aware Attention (CSAA).
The word encoder returns word vectors with both local and global contexts to cover both short and long range dependency relationship between words.
The CSAA then incorporates user and product information to the contextualized words through an attention mechanism that considers the possible existence of cold-start problems.
The full architecture of the model is presented in Figure \ref{fig:model}. We describe the subparts of the model below.

\subsection{Hybrid contextualized word encoder}

The base model is a word encoder that transforms vectors of words $\{w_i\}$ in the text to new word vectors.
In this paper, we present a fast yet very effective word encoder based on two different off-the-shelf classifiers.


The first part of HCWE is based on a CNN model which is widely used in text classification \cite{Kim2014ConvolutionalNN}. This encoder contextualizes words based on local context words to capture short range relationships between words.
Specifically, we do the convolution operation using filter matrices $W_f \in \mathbb{R}^{h \times d}$ with filter size $h$ to a window of $h$ words. We do this for different sizes of $h$. This produces new feature vectors $c_{i,h}$ as shown below, where $f(.)$ is a non-linear function:
\begin{equation*}
    c_{i,h} = f([w_{i-(h-1)/2}; ...; w_{i+(h-1)/2}]^\top W_f + b_f) \nonumber
\end{equation*}

The convolution operation reduces the number of words differently depending on the filter size $h$. To prevent loss of information and to produce the same amount of feature vectors $c_{i,h}$, we pad the texts dynamically such that when the filter size is $h$, the number of paddings on each side is $(h-1)/2$.
This requires the filter sizes to be odd numbers.
Finally, we concatenate all feature vectors of different $h$'s for each $i$ as the new word vector:
\begin{equation*}
    w_{cnn_i} = [c_{i,h_1}; c_{i, h_2}; ...] \nonumber
\end{equation*}


The second part of HCWE is based on an RNN model which is used when texts are longer and include word dependencies that may not be captured by the CNN model. Specifically, we use a bidirectional LSTM and concatenate the forward and backward hidden state vectors as the new word vector, as shown below:
\begin{align*}
    \overrightarrow{h}_i &= LSTM(w_i, \overrightarrow{h}_{i-1}) \\
    \overleftarrow{h}_i &= LSTM(w_i, \overleftarrow{h}_{i+1}) \\
    w_{rnn_i} &= [\overrightarrow{h}_i; \overleftarrow{h}_i] \nonumber
\end{align*}


The answer to the question whether to use local or global context to encode words for sentiment classification is still unclear, and both CNN and RNN models have previous empirical evidence that they perform better than the other \cite{Kim2014ConvolutionalNN,mccann2017learned}. We believe that both short and long range relationships, captured by CNN and RNN respectively, are useful for sentiment classification. There are already previous attempts to intricately combine both CNN and RNN \cite{zhou2016text}, resulting to a slower model. On the other hand, HCWE resorts to combine them by simply concatenating the word vectors encoded from both CNN and RNN encoders, i.e. $w_i = [w_{cnn_i}; w_{rnn_i}]$. This straightforward yet fast alternative outputs a word vector with semantics contextualized from both local and global contexts. Moreover, they perform as well as complex hierarchical structured models \cite{yang2016hierarchical,chen2016neural} which train very slow.

\subsection{Cold-start aware attention}

Incorporating the user and product information of the text as context vectors $u$ and $p$ to attentively pool the word vectors, i.e. $e(w_i, u, p) = v^\top tanh(W_w w_i + W_u u + W_p p + b)$, has been proven to improve the performance of sentiment classifiers \cite{chen2016neural}. However, this method assumes that the user and product vectors are always present. This is not the case in real world settings where a user/product may be new and has just got its first review. In this case, the vectors $u$ and $p$ are rendered useless and may also contain noisy signals that decrease the overall performance of the models.

To this end, we present an attention mechanism called Cold-Start Aware Attention (CSAA). CSAA operates on the idea that a cold-start user/product can use the information of other similar users/products with sufficient number of reviews. CSAA separates the construction of pooled vectors for user and for product, unlike previous methods that use both user/product information to create a single pooled vector.
Constructing a user/product-specific pooled vector consists of three parts: the distinct pooled vector created using the original user/product, the shared pooled vector created using similar users/products, and the selective gate to select between the distinct and shared vectors. Finally, the user- and product-specific pooled vectors are combined into one final pooled vector.

In the following paragraphs, we discuss the step-by-step process on how the user-specific pooled vector is constructed. A similar process is done to construct the product-specific pooled vector, but is not presented here for conciseness.

The user-specific \textbf{distinct pooled vector} $v^d_u$ is created using a method similar to the additive attention mechanism \cite{Bahdanau2014NeuralMT}, i.e. $v^d_u = att(\{w_i\}, u)$, where the context vector is the distinct vector of user $u$, as shown in the equation below. An equivalent method is used to create the distinct product-specific pooled vector $v^d_p$.
\begin{align*}
    e^d_u(w_i, u) &= v^d{}^\top tanh(W^d_w w_i + W^d_u u + b^d) \\
    a^d_{u_i} &= \frac{exp(e^d_u(w_i, u))}{\sum_j exp(e^d_u(w_j, u))} \\
    v^d_u &= \sum_i a^d_{u_i} \times w_i
\end{align*}
The user-specific \textbf{shared pooled vector} $v^s_u$ is created using the same method above, but using a shared context vector $u'$.
The shared context vector $u'$ is constructed using the vectors of other users and weighted based on a similarity weight. Similarity is defined as how similar the word usages of two users are.
This means that if a user $u_k$ uses words similarly to the word usage of the original user $u$, then $u_k$ receives a high similarity weight.
The similarity weight $a^s_{u_k}$ is calculated as the softmax of the product of $\mu(\{w_i\})$ and $u_k$ with a project matrix in the middle, where $\mu(\{w_i\})$ is the average of the word vectors. The similarity weights are used to create $u'$, as shown below. Similar method is used for the shared product-specific pooled vector $v^s_p$.

\begin{align*}
    e^s_u(\mu(\{w_i\}), u_k) &= \mu(\{w_i\}) W^s_u u_k \\
    a^s_{u_k} &= \frac{exp(e^s_u(w_i, u_k))}{\sum_j exp(e^s_u(w_i, u_j))} \\
    u' &= \sum_k a^s_{u_k} \times u_k \\
    v^s_u &= att(\{w_i\}, u')
\end{align*}

We select between the user-specific distinct and shared pooled vector, $v^d_u$ and $v^s_u$, into one user-specific pooled vector $v_u$ through a gate vector $g_u$. The vector $g_u$ should put more weight to the distinct vector when user $u$ is not cold-start and to the shared vector when $u$ is otherwise. We use a \textbf{frequency-guided selective gate} that utilizes the frequency, i.e. the number of reviews user $u$ has written. The challenge is that we do not know how many reviews should be considered cold-start or not. This is automatically learned through a two-parameter Weibull cumulative distribution where given the review frequency of the user $f(u)$, a learned shape vector $k_u$ and a learned scale vector $\lambda_u$, a probability vector is sampled and is used as the gate vector $g_u$ to create $v_u$, according to the equation below.
We normalized $f(u)$ by dividing it to the average user review frequency.
The relu function ensures that both $k_u$ and $\lambda_u$ are non-negative vectors. The final product-specific pooled vector $v_p$ is created in a similar manner.
\begin{align*}
    g_u &= 1 - exp\bigg(-\Big(\frac{f(u)}{relu(\lambda_u)}\Big)^{relu(k_u)}\bigg) \\
    v_u &= g_u \times v^d_u + (1 - g_u) \times v^s_u
\end{align*}

Finally, we combine both the user- and product-specific pooled vector, $v_u$ and $v_p$, into one pooled vector $v_{up}$. This is done by using a gate vector $g_{up}$ created using a sigmoidal transformation of the concatenation of $v_u$ and $v_p$, as illustrated in the equation below.
\begin{align*}
    g_{up} &= \sigma(W_g [v_u; v_p] + b_g) \\
    v_{up} &= g_{up} \times v_u + (1 - g_{up}) \times v_p
\end{align*}

We note that our attention mechanism can be applied to any word encoders, including the basic bag of words (BoW) to more recent models such as CNN and RNN. Later (in Section \ref{sec:results}), we show that CSAA improves the performance of simpler models greatly.

\subsection{Training objective}

Normally, a sentiment classifier transforms the final vector $v_{up}$, usually in a linear fashion, into a vector with a dimension equivalent to the number of classes $C$. A softmax layer is then used to obtain a probability distribution $y'$ over the sentiment classes. Finally, the full model uses a cross-entropy over all training documents $D$ as objective function $L$ during training, where $y$ is the gold probability distribution:
\begin{align*}
    y' &= softmax(W v_{up} + b) \\
    L &= - \sum_{d \in D} \sum_{c \in C} y^{(d)}_c \cdot \log(y'^{(d)}_c)
\end{align*}

However, HCSC has a nice architecture which can be used to improve the training.
It contains seven pooled vectors $\mathbb{V} = \{v^d_u, v^d_p, v^s_u, v^s_p, v_u, v_p, v_{up}\}$ that are essentially in the same vector space. This is because these vectors are created using weighted sums of either the encoded word vectors through attention or the parent pooled vectors through the selective gates. Therefore, we can train separate classifiers for each pooled vectors using the same parameters $W$ and $b$. Specifically, for each $v \in \mathbb{V}$, we calculate the loss $L_v$ using the above formulas. The final loss is then the sum of all the losses, i.e. $L = \sum_{v \in \mathbb{V}} L_v$.

\section{Experiments}

In this section, we present our experiments and the corresponding results. We use the models described in Section \ref{sec:rel} as baseline models: \textbf{JMARS} \cite{diao2014jointly}, \textbf{UPNN} \cite{tang2015learning}, \textbf{TLFM+PRC} \cite{song2017recommendation}, \textbf{UPDMN} \cite{dou2017capturing}, \textbf{TUPCNN} \cite{chen2016learning}, and \textbf{NSC} \cite{chen2016neural}, where \textbf{NSC} is the model with state-of-the-art results.

\subsection{Experimental settings}

\begin{table*}[t]
  \small
  \centering
    \begin{tabular}{|c|c|ccc|ccc|ccc|}
    \hline
    \multirow{2}[1]{*}{Datasets} & \multirow{2}[1]{*}{Classes} & \multicolumn{3}{c|}{Train} & \multicolumn{3}{c|}{Dev} & \multicolumn{3}{c|}{Test} \\
\cline{3-11}     &  & \#docs & \#users & \#prods & \#docs & \#users & \#prods & \#docs & \#users & \#prods \\
    \hline
    IMDB & 10 & 67426 & 1310 & 1635 & 8381 & 1310 & 1574 & 9112 & 1310 & 1578 \\
    Yelp 2013 & 5 & 62522 & 1631 & 1633 & 7773 & 1631 & 1559 & 8671 & 1631 & 1577 \\
    \hline
    \end{tabular}
    
    \begin{tabular}{|c|c|ccc|ccc|ccc|}
    \hline
    \multirow{2}[1]{*}{Datasets} & \multirow{2}[1]{*}{Classes} & \multicolumn{3}{c|}{Sparse20} & \multicolumn{3}{c|}{Sparse50} & \multicolumn{3}{c|}{Sparse80} \\
\cline{3-11}     &  & \#docs & \#users & \#prods & \#docs & \#users & \#prods & \#docs & \#users & \#prods \\
    \hline
    IMDB & 10 & 44261 & 1042 & 1323 & 17963 & 659 & 840 & 2450 & 250 & 312 \\
    Yelp 2013 & 5 & 38687 & 1301 & 1288 & 16058 & 818 & 823 & 2406 & 352 & 304 \\
    \hline
    \end{tabular}%
  \caption{Dataset statistics}
  \label{tab:data}%
\end{table*}%

\paragraph{Implementation}

We set the size of the word, user, and product vectors to 300 dimensions. We use pre-trained GloVe embeddings\footnote{\url{https://nlp.stanford.edu/projects/glove/}} \cite{pennington2014glove} to initialize our word vectors.
We simply set the parameters for both BiLSTMs and CNN to produce an output with 300 dimensions:
For the BiLSTMs, we set the state sizes of the LSTMs to 75 dimensions, for a total of 150 dimensions. For CNN, we set $h=3, 5, 7$, each with 50 feature maps, for a total of 150 dimensions.
These two are concatenated to create a 300-dimension encoded word vectors. We use dropout \cite{srivastava2014dropout} on all non-linear connections with a dropout rate of 0.5. 
We set the batch size to 32.
Training is done via stochastic gradient descent over shuffled mini-batches with the Adadelta update rule \cite{Zeiler2012ADADELTAAA}, with $l_2$ constraint \cite{Hinton2012ImprovingNN} of 3. We perform early stopping using a subset of the given development dataset. Training and experiments are all done using a NVIDIA GeForce GTX 1080 Ti graphics card.

Additionally, we also implement two versions of our model where the word encoder is a subpart of HCSC, i.e. (a) the CNN-based model (\textbf{CNN+CSAA}) and (b) the RNN-based model (\textbf{RNN+CSAA}). For the CNN-based model, we use 100 feature maps for each of the filter sizes $h=3, 5, 7$, for a total of 300 dimensions. For the RNN-based model, we set the state sizes of the LSTMs to 150, for a total of 300 dimensions.

\paragraph{Datasets and evaluation}

We evaluate and compare our models with other competing models using two widely used sentiment classification datasets with available user and product information: IMDB and Yelp 2013. Both datasets are curated by \citet{tang2015learning}, where they are divided into train, dev, and test sets using a 8:1:1 ratio, and are tokenized and sentence-splitted using Stanford CoreNLP \cite{manning2014stanford}. In addition, we create three subsets of the train dataset to test the robustness of the models on sparse datasets. To create these datasets, we randomly remove all the reviews of $x\%$ of all users and products, where $x=20,50,80$. These datasets are not only more sparse than the original datasets, but also have smaller number of users and products, introducing cold-start users and products. All datasets are summarized in Table \ref{tab:data}.
Evaluation is done using two metrics: the Accuracy which measures the overall sentiment classification performance and the RMSE which measures the divergence between predicted and ground truth classes.
We notice very minimal differences among performances of different runs.

\subsection{Comparisons on original datasets}
\label{sec:results}

\begin{table}[t]
  \small
  \centering
    \begin{tabular}{|c|cc|cc|}
    \hline
    \multirow{2}[1]{*}{Models} & \multicolumn{2}{c|}{IMDB} & \multicolumn{2}{c|}{Yelp 2013} \\
\cline{2-5}      & Acc. & RMSE & Acc. & RMSE \\
    \hline
    JMARS & - & 1.773${}^*$ & - & 0.985${}^*$ \\
    UPNN & 0.435${}^*$ & 1.602${}^*$ & 0.596${}^*$ & 0.784${}^*$ \\
    TLFM+PRC & - & 1.352${}^*$ & - & 0.716${}^*$ \\
    UPDMN & 0.465${}^*$ & 1.351${}^*$ & 0.639${}^*$ & 0.662 \\
    TUPCNN & 0.488${}^*$ & 1.451${}^*$ & 0.639${}^*$ & 0.694${}^*$ \\
    NSC & 0.533 & 1.281${}^*$ & 0.650 & 0.692${}^*$ \\
    \hline
    CNN+CSAA & 0.522${}^*$ & 1.256${}^*$ & 0.654 & 0.665 \\
    RNN+CSAA & 0.527${}^*$ & 1.237${}^*$ & 0.654 & 0.667 \\
    \textbf{HCSC} & \textbf{0.542} & \textbf{1.213} & \textbf{0.657} & \textbf{0.660} \\
    \hline
    \end{tabular}%
  \caption{Accuracy and RMSE values of competing models on the original non-sparse datasets. An asterisk indicates that HCSC is significantly better than the model ($p < 0.01$).}
  \label{tab:res_orig}%
\end{table}%

We report the results on the original datasets in Table \ref{tab:res_orig}. On both datasets, HCSC outperforms all previous models based on both accuracy and RMSE. Based on accuracy, HCSC performs significantly better than all previous models except NSC, where it performs slightly better with 0.9\% and 0.7\% increase on IMDB and Yelp 2013 datasets. Based on RMSE, HCSC performs significantly better than all previous models, except when compared with UPDMN on the Yelp 2013 datasets, where it performs slightly better. We note that RMSE is a better metric because it measures how close the wrongly predicted sentiment and the ground truth sentiment are. Although NSC performs as well as HCSC based on accuracy, it performs worse based on RMSE, which means that its predictions deviate far from the original sentiment.

It is also interesting to note that when CSAA is used as attentive pooling, both simple CNN and RNN models perform just as well as NSC, despite NSC being very complex and modeling the documents with compositionality \cite{chen2016neural}. This is especially true when compared using RMSE, where both CNN+CSAA and RNN+CSAA perform significantly better ($p < 0.01$) than NSC. This proves that CSAA is an effective use of the user and product information for sentiment classification.

\subsection{Comparisons on sparse datasets}

\begin{table}[t]
  \small
  \begin{subtable}{0.5\textwidth}
    \centering
    \begin{tabular}{|c|ccc|}
    \hline
    Models & Sparse20 & Sparse50 & Sparse80 \\
    \hline
    NSC(LA) & 0.469 & 0.428 & 0.309  \\
    NSC & 0.497 & \textcolor[rgb]{ 1,  0,  0}{0.408} & \textcolor[rgb]{ 1,  0,  0}{0.292} \\
    \hline
    CNN+CSAA & 0.497 & 0.444 & 0.343 \\
    RNN+CSAA & \textbf{0.505} & 0.455 & 0.364  \\
    \textbf{HCSC} & \textbf{0.505} & \textbf{0.456} & \textbf{0.368}   \\
    \hline
    \end{tabular}%
    \caption{IMDB Datasets}
  \end{subtable}
  \begin{subtable}{0.5\textwidth}
    \centering
    \begin{tabular}{|c|ccc|}
    \hline
    Models & Sparse20 & Sparse50 & Sparse80\\
    \hline
    NSC(LA) & 0.624 & 0.590 & 0.523 \\
    NSC & 0.626 & 0.592 & \textcolor[rgb]{ 1,  0,  0}{0.511} \\
    \hline
    CNN+CSAA & 0.626 & 0.605 & 0.522  \\
    RNN+CSAA & 0.633 & 0.603 & 0.527  \\
    \textbf{HCSC} & \textbf{0.636} & \textbf{0.608} & \textbf{0.538}  \\
    \hline
    \end{tabular}%
    \caption{Yelp 2013 Datasets}
  \end{subtable}
  \caption{Accuracy values of competing models when the training data used is sparse. \textbf{Bold-faced} values are the best accuracies in the column, while \textcolor{red}{red} values are accuracies worse than NSC(LA).}
  \label{tab:res_sparse}%
\end{table}%

Table \ref{tab:res_sparse} shows the accuracy of NSC \cite{chen2016neural} and our models CNN+CSAA, RNN+CSAA, and HCSC on the sparse datasets. As shown in the table, on all datasets with different levels of sparsity, HCSC performs the best among the competing models. The difference between the accuracy of HCSC and NSC increases as the level of sparsity intensifies: While the HCSC only gains 0.8\% and 1.0\% over NSC on the less sparse Sparse20 IMDB and Yelp 2013 datasets, it improves over NSC significantly with 7.6\% and 2.7\% increase on the more sparse Sparse80 IMDB and Yelp 2013 datasets, respectively.

We also run our experiments using NSC without user and product information, i.e. NSC(LA) which reduces the model into a hierarchical LSTM model \cite{yang2016hierarchical}. Results show that although the use of user and product information in NSC improves the model on less sparse datasets (as also shown in the original paper \cite{chen2016neural}), it decreases the performance of the model on more sparse datasets: It performs 2.0\%, 1.7\%, and 1.2\% worse than NSC(LA) on Sparse50 IMDB, Sparse80 IMDB, and Sparse80 Yelp 2013 datasets.
We argue that this is because NSC does not consider the existence of cold-start problems, which makes the additional user and product information more noisy than helpful.

\section{Analysis}

In this section, we show further interesting analyses of the properties of HCSC. We use the Sparse50 datasets and the corresponding results of several models as the experimental data.

\paragraph{Performance per review frequency}

\begin{figure}[t]
    \centering
    \includegraphics[width=0.47\textwidth]{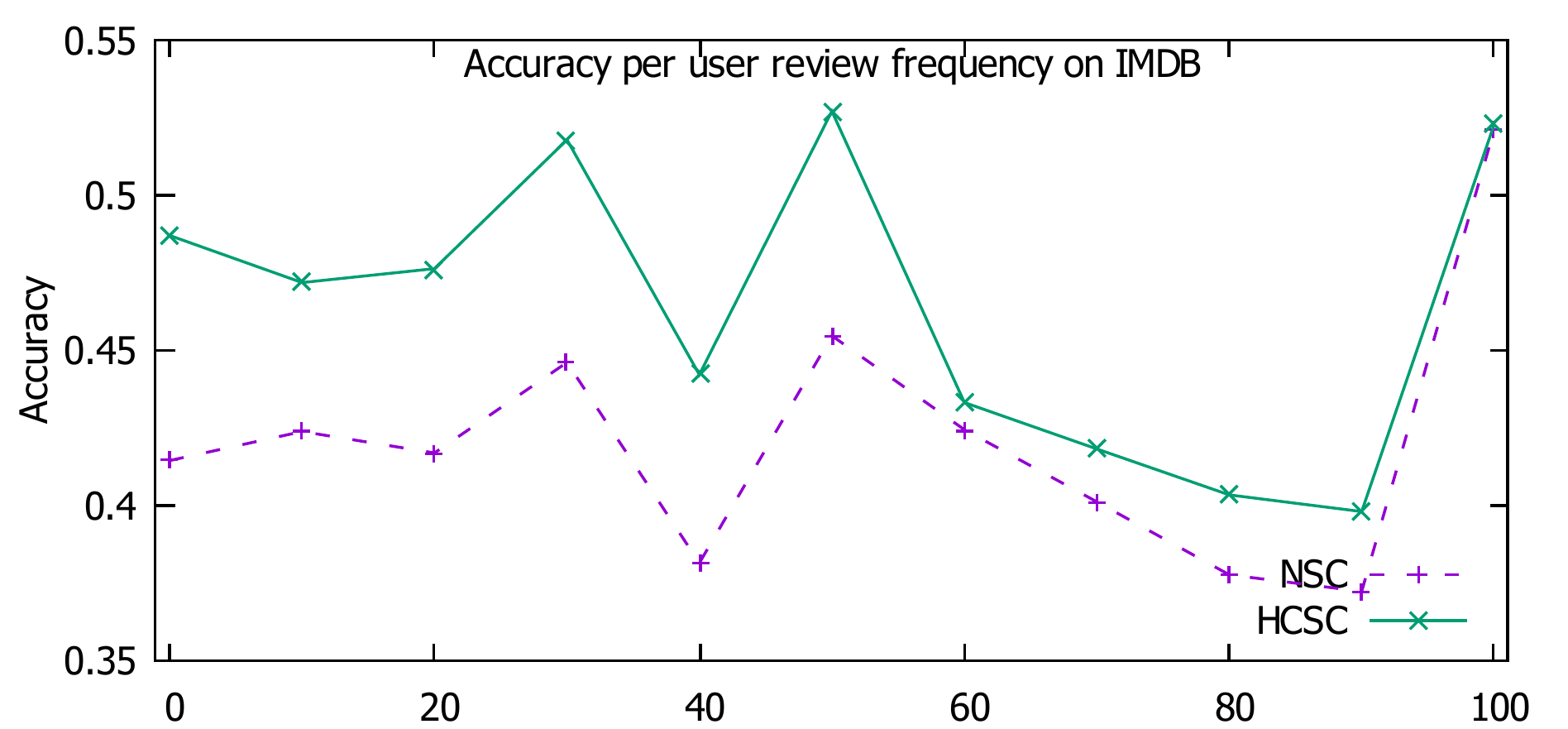}
    \includegraphics[width=0.47\textwidth]{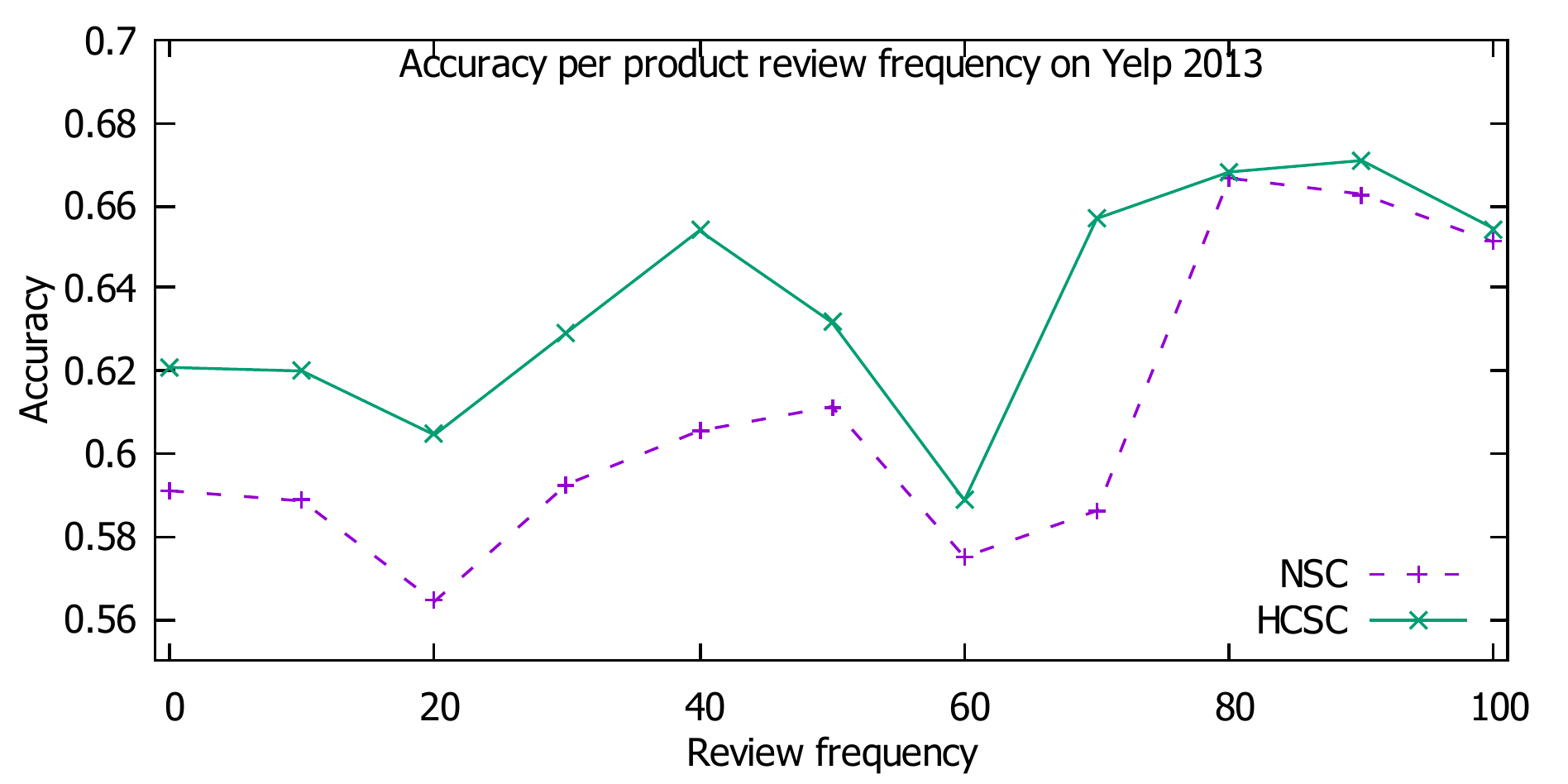}
    \caption{Accuracy per user/product review frequency on both datasets. The review frequency value $f$ represents the frequencies in the range $[f, f+10)$, except when $f=100$, which represents the frequencies in the range $[f, \infty)$.}
    \label{fig:freq_acc}
\end{figure}

We investigate the performance of the model over users/products with different number of reviews. Figure \ref{fig:freq_acc} shows plots of accuracy of both NSC and HCSC over (a) different user review frequency on IMDB dataset and (b) different product review frequency on Yelp 2013 dataset. On both plots, we observe that when the review frequency is small, the performance gain of HCSC over NSC is very large. However, as the review frequency becomes larger, the performance gain of HCSC over NSC decreases to a very marginal increase. This means that HCSC finds its improvements over NSC from cold-start users and products, in which NSC does not consider explicitly.

\paragraph{How few is cold-start?}

\begin{figure*}[t]
    \centering
    \includegraphics[width=0.97\textwidth]{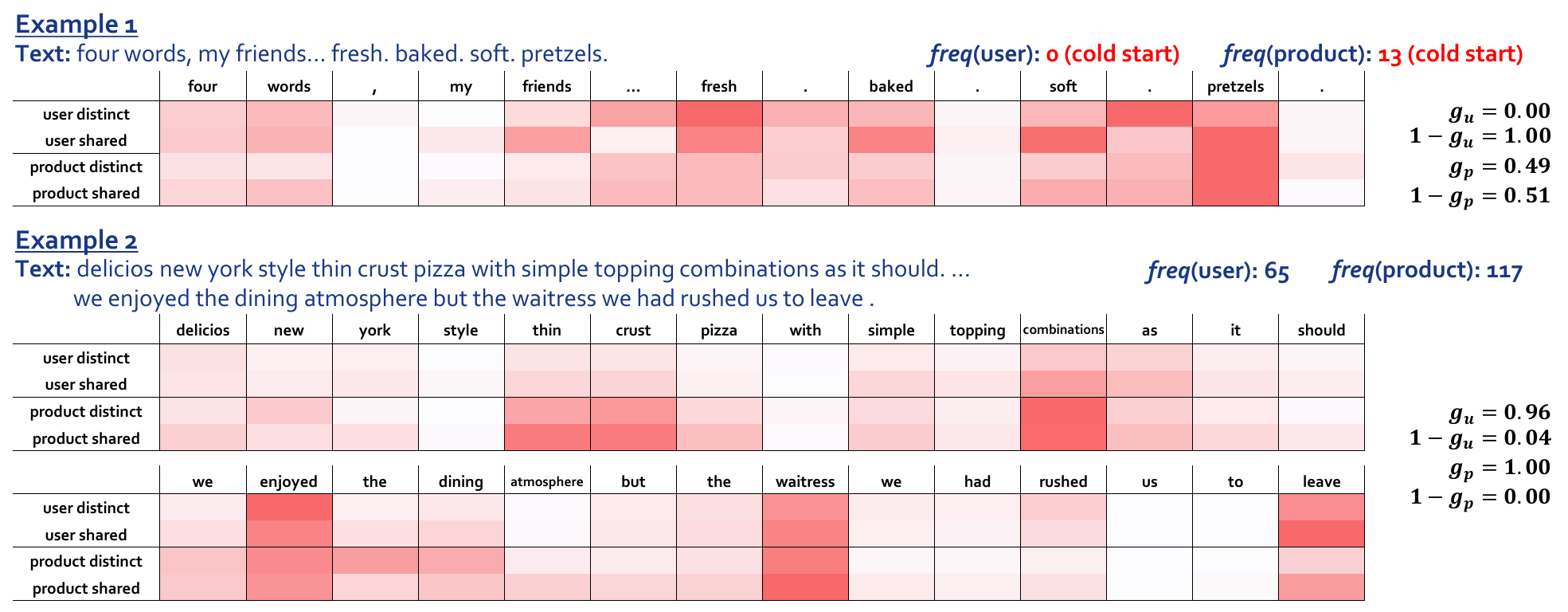}
    \caption{Visualization of attention and gate values of two examples from the Yelp 2013 dataset. Example 2 is truncated, leaving only the important parts. Gate values $g$'s are the average of the values in the original gate vector.}
    \label{fig:examples}
\end{figure*}

\begin{figure}[t]
    \centering
    \includegraphics[width=0.47\textwidth]{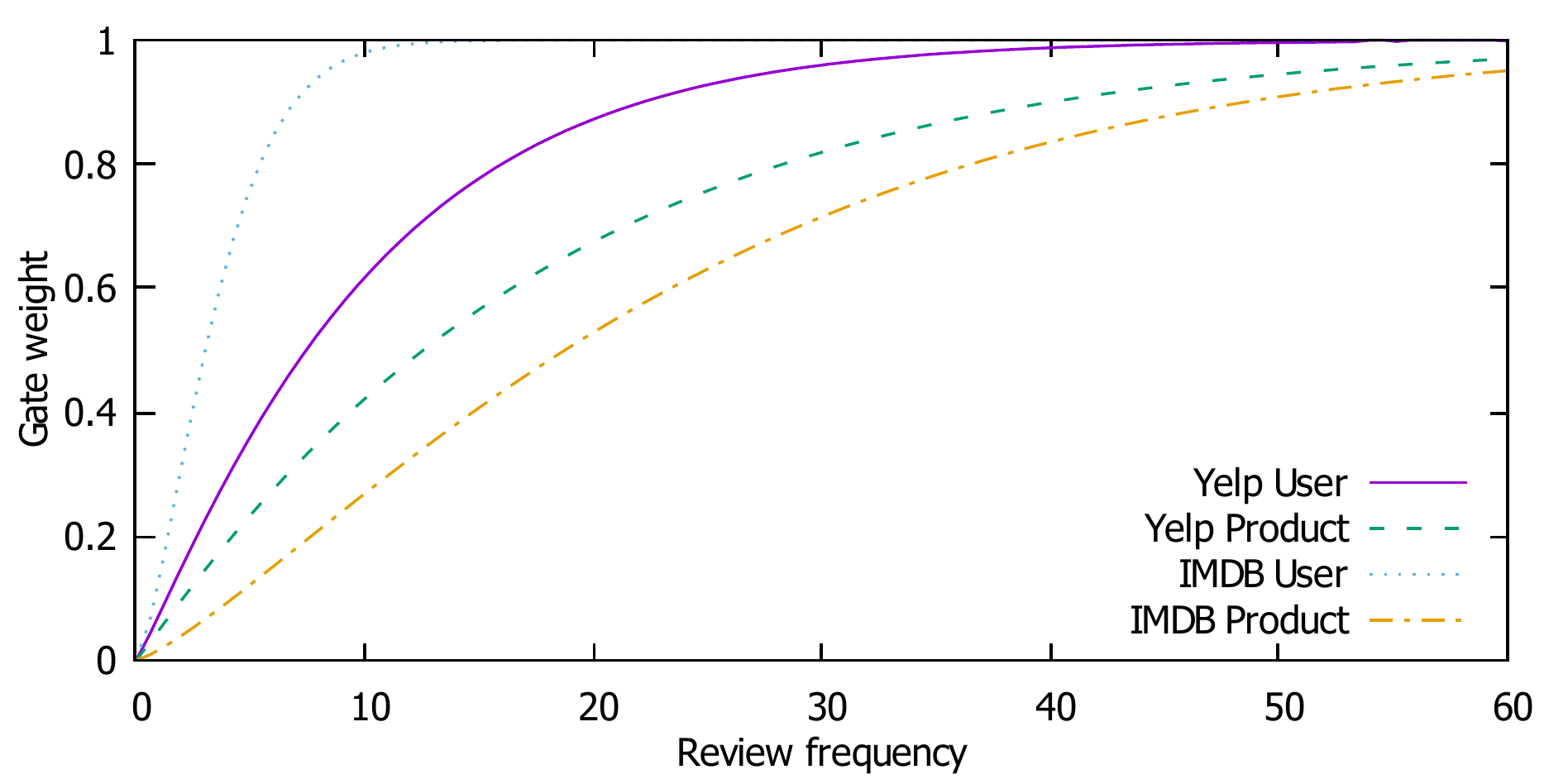}
    \caption{Graph of the user/product-specific Weibull cumulative distribution on both datasets.}
    \label{fig:weibull}
\end{figure}

One intriguing question is when do we say that a user/product is cold-start or not. Obviously, users/products with no previous reviews at all should be considered cold-start, however the cut-off point between cold-start and non-cold-start entities is vague. Although we cannot provide an exact answer to this question, HCSC is able to provide a nice visualization by reducing the shape and scale vectors, $k$ and $\lambda$, of the frequency-guided selective gate into their averages and draw a Weibull cumulative distribution graph, as shown in Figure \ref{fig:weibull}. The figure provides us these observations: First, users have a more lenient cold-start cut-off point compared to products; in the IMDB dataset, a user only needs approximately at least five reviews to use at least 80\% of its own information (i.e. distinct vector). On the other hand, products tend to need more reviews to be considered sufficient and not cold start; in the IMDB dataset, a product needs approximately 40 reviews to use at least 80\% of its own information. This explains the marginal increase in performance of previous models when only product information is used as additional context, as reported by previous papers \cite{tang2015learning,chen2016neural}.

\paragraph{On the different pooled vectors}

We visualize the attention and gate values of two example results from HCSC in Figure \ref{fig:examples} to investigate on how user/product vectors, and distinct/shared vectors work.
In the first example, both user and product are cold-start. The user distinct vector focuses its attention to wrong words, since it is not able to use any useful information from the user at all. In this case, HCSC uses the user shared vector by using a gate vector $g_u=0$. The user shared vector correctly attends to important words such as \textit{fresh}, \textit{baked}, \textit{soft}, and \textit{pretzels}.
In the second example, both user and product are not cold-start. In this case, the distinct vectors are used almost entirely by setting the gates close to 1. Still, the corresponding shared vectors are similar to the distinct vectors, proving that HCSC is able to create useful user/product-specific context from similar users/products.
Finally, we look at the differing attention values of users and products. We observe that user vectors focus on words that describe the product or express their emotions (e.g. \textit{fresh} and \textit{enjoyed}). On the other hand, product vectors focus more on words pertaining to the products/services (e.g. \textit{pretzels} and \textit{waitress}).

\paragraph{On the time complexity of models}

\begin{table}[t]
  \small
  \centering
    \begin{tabular}{|c|cc|}
    \hline
    Models & IMDB & Yelp 2013 \\
    \hline
    NSC & 7331 & 6569 \\
    \hline
    CNN+CSAA & 256 (28.6x) & 146 (45.0x) \\
    RNN+CSAA & 968 (7.6x) & 561 (11.7x) \\
    HCSC & 1110 (6.6x) & 615 (10.7x) \\
    \hline
    \end{tabular}%
  \caption{Time (in seconds) to process the first 100 batches of competing models for each dataset. The numbers in the parenthesis are the speedup of time when compared to NSC.}
  \label{tab:time}%
\end{table}%

Finally, we report the time in seconds to run 100 batches of data of the models NSC, CNN+CSAA, RNN+CSAA, and HCSC in Figure \ref{tab:time}. NSC takes too long to train, needing at least 6500 seconds to process 100 batches of data. This is because it uses two non-parallelizable LSTMs on top of each other. Our models, on the other hand, only use one (or none in the case of CNN+CSAA) level of BiLSTM. This results to at least 6.6x speedup on the IMDB datasets, and at least 10.7x speedup on the Yelp 2013 datasets. This means that HCSC does not sacrifice a lot of time complexity to obtain better results.

\section{Conclusion}

We propose Hybrid Contextualized Sentiment Classifier (HCSC) with a fast word encoder which contextualizes words to contain both short and long range word dependency features, and an attention mechanism called Cold-start Aware Attention (CSAA) which considers the existence of the cold-start problem among users and products by using a shared vector and a frequency-guided selective gate, in addition to the original distinct vector. Our experimental results show that our model performs significantly better than previous models. These improvements increase when the level of sparsity in data increases, which confirm that HCSC is able to deal with the cold-start problem.

\section*{Acknowledgements}

This work was supported by Microsoft Research Asia and the ICT R\&D program of MSIT/IITP.
[2017-0-01778, Development of Explainable Human-level Deep Machine Learning Inference Framework]

\bibliography{acl2018}
\bibliographystyle{acl_natbib}

\end{document}